\documentclass[sn-basic]{sn-jnl}

\usepackage{amsmath}     
\usepackage{manyfoot}    
\usepackage{booktabs}
\usepackage{array}
\usepackage{siunitx}
\usepackage{xcolor}
\usepackage{tikz}
\usetikzlibrary{shapes.geometric, arrows.meta, positioning, fit, backgrounds}

\AtBeginDocument{\setlength{\bibsep}{0.8em}}

\begin{document}

\title[Towards Automated Exam Grading]{Towards Fully Automated Exam Grading:
Fairness-Aware Recognition of Handwritten Answers with Foundation Models}

\author[1]{\fnm{Hartwig} \sur{Grabowski}
  (\href{https://orcid.org/0009-0001-4300-2626}{ORCID: 0009-0001-4300-2626})}%
  \email{hartwig.grabowski@hs-offenburg.de}

\affil[1]{\orgdiv{Institute for Machine Learning and Analytics (IMLA)},
\orgname{Offenburg University}, \city{Offenburg}, \country{Germany}}

\abstract{Correcting handwritten exams by hand is time-consuming and
error-prone, particularly for large cohorts, while fully digital exams tend to
force a didactic narrowing towards closed question formats. A practical middle
ground keeps paper-based, problem-oriented tasks but records the
assessment-relevant answers as single capital letters in a table that a machine
can read. The open question is whether this reading can be made accurate and,
above all, \emph{fair} enough for unsupervised grading. Earlier automated
approaches reached only about \SI{88}{\percent}--\SI{91}{\percent} recognition
--- too low --- and failed on the cases that matter most: answers placed
outside the cell, crossed out, or written in cursive. We show that
general-purpose \emph{vision--language foundation models} (VLMs), which
interpret the page rather than match pixel templates, close this gap. On a
benchmark of 61 anonymised exams (\num{3141} answer positions) the best model
reaches \SI{98.4}{\percent} accuracy, well above the previous baseline.
Crucially, we centre the evaluation on \emph{fairness}: we distinguish false
negatives (a correct answer marked wrong, which disadvantages the student) from
false positives, and a lightweight prompt that supplies the reference solution
as context lowers the false-negative rate to \SI{0.58}{\percent}. Under an
exemplary grading scheme only three of the 61 exams would be graded worse, all
caught by a student self-review step. Fully automated, fairness-aware exam
grading at scale is therefore defensible; we release the anonymised benchmark
to support reproducibility.}


\maketitle

\section{Introduction}
Paper-based exams remain common, and automating their correction can save
substantial time while reducing transcription errors. We consider the
use-case introduced in our previous work~\cite{grabowski2023icr}: a printed
table is placed in the lower-right area of each exam page; each column
corresponds to one question and is answered by a single capital Latin letter
(``A'',\,\ldots,\,``Z''). Recognising these letters automatically turns the
exam into machine-gradable data.

This coded answer format is a deliberate design choice: the exam stays on
paper, so instructors keep full freedom to pose multi-step, problem-solving
tasks, while only the assessment-relevant partial results are entered as single
capital letters into table cells. Fully digital summative exams, in contrast,
tend to force a narrowing towards closed (e.g.\ multiple-choice)
formats~\cite{grabowski2026hybrid}. The coded format therefore enables
automated grading \emph{without} constraining task design --- provided the
handwritten letters can be recognised reliably, which is exactly the technical
problem we study here.

In~\cite{grabowski2023icr} we developed a sequence of detector-based
pipelines, culminating in a YOLOv5 model trained on synthetically generated
forms, reaching an overall recognition rate of \SI{88.28}{\percent}. We
concluded that this is \emph{not} sufficient for fully automated grading and
proposed an interactive system with human verification and an e-mail
self-check for students. The present paper closes that gap.

Our central claim is that general-purpose \emph{vision--language foundation
models} (VLMs), i.e.\ broadly pre-trained models adapted to many downstream
tasks~\cite{bommasani2021foundation}, are now strong enough to make fully
automated correction defensible. Concretely, we contribute:
\begin{enumerate}
  \item A \textbf{unified benchmark} (61 anonymised exams, \num{3141} answer
  positions) on which we compare five current frontier VLMs against the YOLOv5
  / YOLO26 baselines under identical evaluation rules.
  \item A \textbf{fairness-aware evaluation} that distinguishes
  false negatives (FN, student disadvantaged) from false positives (FP,
  student over-credited), and a simple \emph{weak-hint} prompting scheme that
  drives the FN rate below \SI{1}{\percent}.
  \item A \textbf{practicability analysis}: combining the low FN rate with
  student self-review, we estimate that only a small number of exams would be
  affected at the grade level, supporting fully automated correction.
  \item An \textbf{end-to-end, unattended pipeline} (scan $\rightarrow$
  recognise $\rightarrow$ score $\rightarrow$ grade $\rightarrow$ notify) whose
  only manual step is an exception-driven student complaint check, making the
  correction of \emph{large} paper-based cohorts feasible --- a regime where
  neither fully digital exams nor manual tutor grading scale well.
\end{enumerate}

\section{Background and Related Work}
\paragraph{Detector-based recognition (prior work).}
Our earlier study~\cite{grabowski2023icr} compared four approaches:
fixed-cell cropping with a CNN classifier (\SI{70.75}{\percent}), YOLOv5-based
cell cropping with a CNN (\SI{65.50}{\percent}), YOLOv5 for joint
segmentation and classification trained on EMNIST~\cite{cohen2017emnist}
(\SI{84.46}{\percent}), and the same architecture trained on a customised
dataset (\SI{88.28}{\percent}). The analysis revealed structural limits of
the EMNIST label space (e.g.\ ``O''/``0'' and ``I''/``L'' confusions) and,
more importantly, error classes that are hard or impossible to train for: a
detector trained on block capitals inside cells cannot reliably handle
letters placed \emph{above/below} the table, \emph{crossed-out} answers, or
answers in \emph{cursive}.

\paragraph{Handwritten text recognition.}
Handwritten Text Recognition (HTR) has developed from word- and line-level
recognition systems towards document-level neural approaches; a recent survey
summarises this progression, the relevant datasets, and the remaining
robustness challenges caused by handwriting variability~\cite{garrido2026htrsurvey}.
Our task is narrower than general HTR because each answer position contains a
single capital letter, but it is also more decision-critical: a one-character
recognition error directly changes awarded points.

\paragraph{Pre-trained OCR and document models.}
Recent OCR systems increasingly use pre-trained Transformer components rather
than task-specific CNN/RNN pipelines. TrOCR combines pre-trained image and
text Transformers and reports strong results on printed, scene, and
handwritten text recognition~\cite{li2023trocr}. For full-page document
understanding, Donut removes the external OCR stage entirely and trains an
OCR-free encoder--decoder model for visual document understanding
tasks~\cite{kim2022donut}; Nougat similarly casts scientific-document
conversion as an end-to-end visual Transformer problem~\cite{blecher2023nougat}.
These systems motivate our move away from isolated glyph detectors, but they
are still specialised architectures rather than general interactive VLMs used
directly on exam pages.

\paragraph{Vision--language foundation models.}
Large multimodal models combine a vision encoder with a language model and
perform OCR / document understanding zero-shot, exploiting world knowledge
and context rather than a fixed class vocabulary. General VLM architectures
such as BLIP-2 and LLaVA connect frozen or instruction-tuned language models
to visual encoders~\cite{li2023blip2,liu2023llava}, while Qwen-VL explicitly
targets text reading and grounding in addition to generic vision--language
tasks~\cite{bai2023qwen}. Benchmarks such as OCRBench and CC-OCR show that
OCR remains a non-trivial capability for large multimodal models, especially
for handwritten and fine-grained text-related
tasks~\cite{liu2024ocrbench,yang2024ccocr}. Work
on online handwriting further indicates that VLMs need suitable
representations to handle handwriting reliably~\cite{fadeeva2024onlinehandwriting}.
Recent HTR benchmarks for multimodal LLMs likewise report strong potential,
especially for modern handwriting, but also show that performance depends on
language, document type, and prompting~\cite{crosilla2025llmhtr}.
We evaluate five current frontier models: Google Gemini~3.1~Flash-Lite and
3.5~Flash~\cite{gemini31flashlite,gemini35flash}, OpenAI
GPT-5.2~\cite{gpt52}, Alibaba Qwen3-VL~\cite{qwen3vl}, and xAI
Grok-4.3~\cite{grok43}. To our knowledge, their suitability for \emph{fair}
automated exam grading---in particular the trade-off between disadvantaging
and over-crediting students---has not been studied.

\section{Why the Detector Fails: An Error Taxonomy}
\label{sec:failure}
\begin{figure*}[t]
\centering
\includegraphics[width=0.66\linewidth]{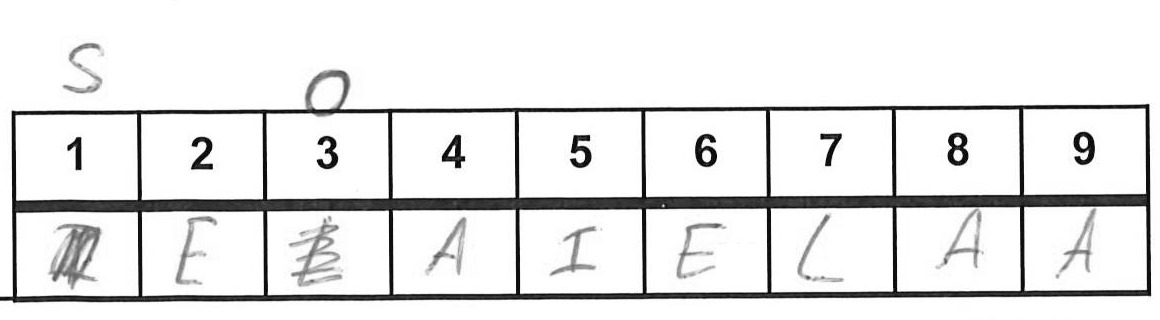}\\[2pt]
{\small (a) out-of-cell placement \& column shift}\\[6pt]
\begin{tabular}{@{}c@{\hskip 10pt}c@{\hskip 10pt}c@{}}
\includegraphics[width=0.21\linewidth]{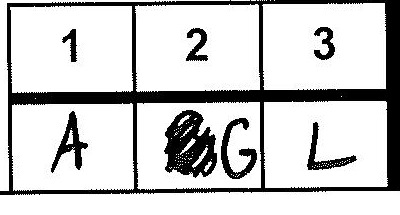} &
\includegraphics[width=0.21\linewidth]{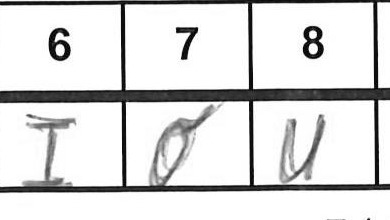} &
\includegraphics[width=0.21\linewidth]{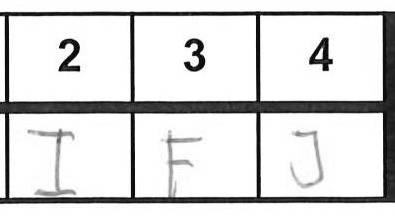} \\
{\small (b) crossed-out + valid ``G''} & {\small (c) cursive ``O''} & {\small (d) ``F'' on cell border} \\
\end{tabular}
\caption{Representative detector failures that the VLM recovers.
(a)~A letter written \emph{above} column~3 (the ``O'') is mis-assigned, shifting
every following entry. (b)~Cell~2 holds a crossed-out scribble \emph{and} a
valid ``G''; YOLO returns nothing, the VLM reads ``G''. (c)~A cursive ``O''
(not in EMNIST) is missed by YOLO. (d)~An ``F'' whose lower stroke meets the
cell border is read as ``E''.}
\label{fig:failures}
\end{figure*}

To understand \emph{why} the VLMs outperform the detector, we inspected the
\num{229} answer positions on which YOLOv5 is wrong while Gemini~3.1~Flash-Lite
(\texttt{plain}) is correct. In \num{92} of these the detector returned no
letter at all (a missed detection); in \num{137} it returned a wrong letter
(a misclassification). The cases fall into four categories
(Table~\ref{tab:taxonomy}, Fig.~\ref{fig:failures}).

\begin{table*}[t]
\caption{Error taxonomy of detector failures that the VLM recovers, derived
from the \num{229} YOLO-wrong / Gemini-correct positions.}
\label{tab:taxonomy}
\centering
\footnotesize
\renewcommand{\arraystretch}{1.2}
\begin{tabular}{@{}>{\raggedright\arraybackslash}p{0.20\textwidth}>{\raggedright\arraybackslash}p{0.56\textwidth}>{\raggedright\arraybackslash}p{0.16\textwidth}@{}}
\toprule
\textbf{Category} & \textbf{Root cause} & \textbf{Type} \\
\midrule
(1)~Placement \& column assignment & letter written above/below the cell; a mark above one column shifts the whole row's assignment & semantic \\
\midrule
(2)~Crossed-out / multiple marks & a struck-through answer, or a crossed-out \emph{and} a valid letter in the same cell; the intended answer must be inferred & semantic \\
\midrule
(3)~Out-of-distribution glyphs & cursive letters and European print variants (e.g.\ hooked ``G'', round ``L'') absent from EMNIST & distribution \\
\midrule
(4)~Letter--border interaction & ``F'' vs.\ ``E'' when the lower stroke coincides with the cell's bottom line & distribution / training \\
\bottomrule
\end{tabular}
\end{table*}

Categories~(1) and~(2) are \emph{semantic}: deciding which mark is the intended
answer, or to which column a stray letter belongs, requires interpreting the
page layout --- exactly the strength of a vision--language model.
Categories~(3) and~(4) reflect the \emph{distributional} limits of a detector
tied to a fixed class vocabulary: cursive and European letter forms are simply
absent from EMNIST, and the F/E border ambiguity is a training-data placement
artifact --- letters were projected into cells without aligning their baseline
to the cell border, so the detector cannot separate the cell line from the
letter's own stroke. A general VLM, pre-trained on vast and varied text,
sidesteps both failure modes.

\section{Method}
\subsection{Benchmark and Ground Truth}
We use 61 anonymised exams (matriculation numbers and point headers removed),
comprising \num{742} scanned pages and \num{3141} answer positions. For each
exam we have (i) the per-page student answers (\texttt{studSolution}), and
(ii) the official correction (\texttt{result}), which yields, per position,
both the reference letter and whether the student's answer was awarded points.

\subsection{Models and Inference Modes}
We evaluate the YOLO baselines (YOLOv5~Medium~\cite{yolov5},
YOLO26~Medium~\cite{ultralytics}) and five VLMs accessed via their provider
APIs. Each page image is sent with a fixed prompt requesting the letters per
column as structured output. We compare two inference modes:
\begin{itemize}
  \item \textbf{\texttt{plain}}: the model receives only the page image.
  \item \textbf{\texttt{weak-hint}}: the model additionally receives the
  reference solution of that page as a line-aligned \emph{hint} (``the expected
  letters are \ldots''), without being told the student's answer. This is a
  cheap way to bias the model towards reading a genuinely-correct answer as
  correct, i.e.\ to reduce false negatives.
\end{itemize}
VLM runs use the native Google API for Gemini (which we found more robust than
a third-party router for the newest Flash model) and OpenRouter for the
others; model snapshot dates are listed in Table~\ref{tab:main}.

\subsection{Evaluation Metrics}
\paragraph{Token accuracy.} Per line we compare the first $x$ predicted tokens
against the $x$ ground-truth tokens; accuracy is correct\,/\,total over all
\num{3141} positions, reported with a \SI{95}{\percent} Wilson confidence
interval.
\paragraph{Fairness (classification) view.} Treating ``answer is
content-correct'' as the positive class, we compute, per position aligned to
the official correction:
\emph{TP} (correct answer recognised as correct),
\emph{FN} (correct answer scored as wrong --- \textbf{disadvantages} the
student), \emph{FP} (wrong answer scored as correct --- \textbf{over-credits}
the student), and \emph{TN}. We emphasise the
false-negative rate $\mathrm{FNR}=\mathrm{FN}/(\mathrm{TP}+\mathrm{FN})$ as the
key fairness metric and specificity
$\mathrm{TN}/(\mathrm{TN}+\mathrm{FP})$ as the over-crediting / hint-bias
indicator. Significance between models is assessed with a
continuity-corrected McNemar test on the per-position correctness vectors.

\section{Results}
\begin{table*}[t]
\centering
\caption{Recognition on the unified 61-exam benchmark (\num{3141} positions),
grouped per model (\texttt{weak-hint} then \texttt{plain}). Acc.\ with
\SI{95}{\percent} Wilson CI; FP/FN absolute; FNR and specificity in the
fairness view. $^{\dagger}$Gemini via native Google API; others via
OpenRouter.}
\label{tab:main}
\setlength{\tabcolsep}{5pt}
\footnotesize
\renewcommand{\arraystretch}{1.2}
\begin{tabular}{llrrrrrr}
\toprule
Model & Mode & Acc.\,(\%) & 95\,\% CI & FP & FN & FNR\,(\%) & Spec. \\
\midrule
Gemini 3.1 Flash-Lite$^{\dagger}$ & weak-hint & \textbf{98.38} & 97.9--98.8 & 11 & 14 & \textbf{0.58} & 0.985 \\
Gemini 3.1 Flash-Lite$^{\dagger}$ & plain & 97.61 & 97.0--98.1 & 1 & 52 & 2.17 & 0.999 \\
Gemini 3.5 Flash$^{\dagger}$      & weak-hint & 98.22 & 97.7--98.6 & 16 & 12 & 0.50 & 0.979 \\
Gemini 3.5 Flash$^{\dagger}$      & plain & 97.77 & 97.2--98.2 & 1 & 48 & 2.00 & 0.999 \\
Qwen3-VL-30B                      & weak-hint & 96.69 & 96.0--97.3 & 32 & 37 & 1.55 & 0.957 \\
Qwen3-VL-30B                      & plain & 93.92 & 93.0--94.7 & 4 & 137 & 5.72 & 0.995 \\
OpenAI GPT-5.2                    & weak-hint & 95.57 & 94.8--96.2 & 80 & 27 & 1.13 & 0.893 \\
OpenAI GPT-5.2                    & plain & 93.57 & 92.7--94.4 & 5 & 144 & 6.02 & 0.993 \\
xAI Grok-4.3                      & weak-hint & 90.10 & 89.0--91.1 & 193 & 39 & 1.63 & 0.742 \\
xAI Grok-4.3                      & plain & 89.59 & 88.5--90.6 & 1 & 239 & 9.98 & 0.999 \\
\midrule
YOLOv5 Medium                     & plain & 90.93 & 89.9--91.9 & 2 & 183 & 7.64 & 0.997 \\
YOLO26 Medium                     & plain & 89.18 & 88.0--90.2 & 1 & 241 & 10.07 & 0.999 \\
\bottomrule
\end{tabular}
\end{table*}

\paragraph{Foundation models clearly beat the detector baseline.}
Table~\ref{tab:main} shows that the two Gemini Flash models reach
\SI{\sim98}{\percent} token accuracy, roughly \SI{7}{} percentage points above
YOLOv5 (\SI{90.9}{\percent}). The difference is statistically significant
(McNemar, Gemini 3.1 Flash-Lite weak-hint vs.\ YOLOv5: $n_{10}=260$, $n_{01}=26$,
$p=3.5\times10^{-43}$). Qwen3-VL-30B (\SI{96.7}{\percent}) and GPT-5.2
(\SI{95.6}{\percent}) also surpass the baseline; only Grok-4.3 is comparable
to YOLO. Crucially, the VLMs handle the failure cases of
Sec.~\ref{sec:failure}, which the detector cannot.

\paragraph{The newest generation does not help.}
On the full benchmark, Gemini~3.5~Flash (\SI{98.22}{\percent} weak-hint) is
statistically indistinguishable from the cheaper Gemini~3.1~Flash-Lite
(\SI{98.38}{\percent} weak-hint); the confidence intervals overlap. The newest
Flash generation therefore yields no measurable benefit for this task ---
3.1~Flash-Lite is already near the practical ceiling.

\paragraph{Weak hints reduce false negatives; watch the hint bias.}
The \texttt{weak-hint} mode consistently lowers FN relative to \texttt{plain}
(e.g.\ Gemini 3.1 Flash-Lite: $52\to14$; Qwen: $137\to37$; GPT-5.2:
$144\to27$). The cost is an increase in FP for some models --- a
\emph{hint bias} in which the model over-trusts the provided reference and
accepts wrong answers (Grok: FP $1\to193$, specificity $0.74$; GPT-5.2: FP
$5\to80$). Gemini Flash is the only family that keeps \emph{both} FN and FP
low (FP\,=\,11, specificity\,$0.985$), making it the preferred configuration.

\section{Fairness and Practicability}
For fully automated grading the asymmetry of errors matters more than raw
accuracy. An FP over-credits a student and is, at worst, lenient; an FN
\emph{removes} earned points and is the harmful error. We therefore select
\textbf{Gemini~3.1~Flash-Lite in \texttt{weak-hint} mode} as the operating point:
\SI{0.58}{\percent} FNR, i.e.\ \num{14} false negatives among \num{3141}
positions.

These \num{14} FN are distributed over \num{8} of the \num{61} exams. Only
false negatives that actually \emph{change a grade} are practically relevant.
We therefore convert points to grades using an exemplary grading
table\footnote{The 61 exams stem from different courses, so no single grading
table applies in reality. We adopt one representative table and apply it
uniformly to all exams; the resulting counts are thus an estimate of the
expected complaint workload, not an exact institutional figure.} and compare,
per exam, the grade from the true (human) score with the grade an automated
\texttt{weak-hint}-mode correction would assign. The result is summarised in
Table~\ref{tab:grades}, whose columns count exams whose grade is unchanged,
\emph{worse} (lowered $\Rightarrow$ student disadvantaged), or \emph{better}
(raised $\Rightarrow$ over-credited). Of the \num{61} exams, \textbf{only three
receive a worse grade} (the complaint-relevant cases: $2.0\!\to\!2.3$,
$3.7\!\to\!4.0$, $3.0\!\to\!3.3$), while four are slightly over-credited
(student-favourable, no complaint) and 54 are unchanged. In \texttt{plain}
mode the same analysis disadvantages twelve exams --- four times as many ---
which is why we adopt \texttt{weak-hint} as the operating point. Because the corrected
result is returned to each student for self-verification (as already proposed
in~\cite{grabowski2023icr}), the three disadvantaged cases would surface as
complaints and can be reviewed individually --- a negligible workload for 61
exams.

\begin{table}[t]
\centering
\caption{Grade-level effect of fully automated correction (uniform exemplary
grading table).}
\label{tab:grades}
\centering
\footnotesize
\renewcommand{\arraystretch}{1.2}
\begin{tabular}{@{}lccc@{}}
\toprule
Mode & Unchanged & Worse & Better \\
\midrule
Flash-Lite \texttt{weak-hint} & 54 & \textbf{3} & 4 \\
Flash-Lite \texttt{plain} & 49 & 12 & 0 \\
\bottomrule
\end{tabular}
\end{table}

\paragraph{Is a \texttt{plain} consistency check worth it?}
Running both modes and flagging the positions where they disagree realises a
lightweight \emph{four-eyes} (dual-configuration) check that targets
hint-induced FP. For Gemini this would reduce the already-rare FP at the cost
of reviewing dozens of conflicting positions across many pages. Given the low
FP level and the student-favourable direction of FP, we judge this second stage
unnecessary for routine operation, and recommend \texttt{weak-hint}-only with
student self-review.

\section{End-to-End Pipeline and Scalability}
\label{sec:pipeline}
\begin{figure}[!ht]
\centering
\resizebox{0.5\columnwidth}{!}{%
\begin{tikzpicture}[
  node distance=3.5mm, font=\small,
  proc/.style={rectangle, rounded corners, draw, fill=blue!5, align=center,
               text width=58mm, minimum height=6.5mm, inner sep=2pt},
  setup/.style={rectangle, rounded corners, draw, dashed, fill=gray!10,
               align=center, text width=58mm, minimum height=6.5mm, inner sep=2pt},
  manual/.style={rectangle, rounded corners, draw, dashed, fill=orange!15,
               align=center, text width=58mm, minimum height=6.5mm, inner sep=2pt},
  arr/.style={-{Latex}, thick}]
\node[setup] (s0) {\textbf{Setup (once / exam):} reference solution + point table};
\node[proc, below=of s0] (s1) {Scan all exams $\to$ PDF; digitise matriculation no. in separate university-approved OCR step};
\node[proc, below=of s1] (s2) {Completeness check (expected page count / exam)};
\node[proc, below=of s2] (s3) {Auto-crop answer-table regions};
\node[proc, below=of s3] (s4) {VLM letter recognition (\texttt{weak-hint})};
\node[proc, below=of s4] (s5) {Score: letters $\to$ points, sum per exam};
\node[proc, below=of s5] (s6) {Compute grade (grade-point table)};
\node[proc, below=of s6] (s7) {E-mail annotated result to student};
\node[manual, below=of s7] (s8) {\textbf{Manual (exception only):} student self-review --- report \emph{only} on grade worsening};
\foreach \a/\b in {s0/s1,s1/s2,s2/s3,s3/s4,s4/s5,s5/s6,s6/s7,s7/s8} \draw[arr] (\a)--(\b);
\end{tikzpicture}}
\caption{End-to-end correction pipeline. Every step runs unattended and scales
linearly with cohort size; the only human involvement is the exception-driven
complaint check, which the e-mail restricts to grade-worsening (disadvantage)
cases.}
\label{fig:pipeline}
\end{figure}
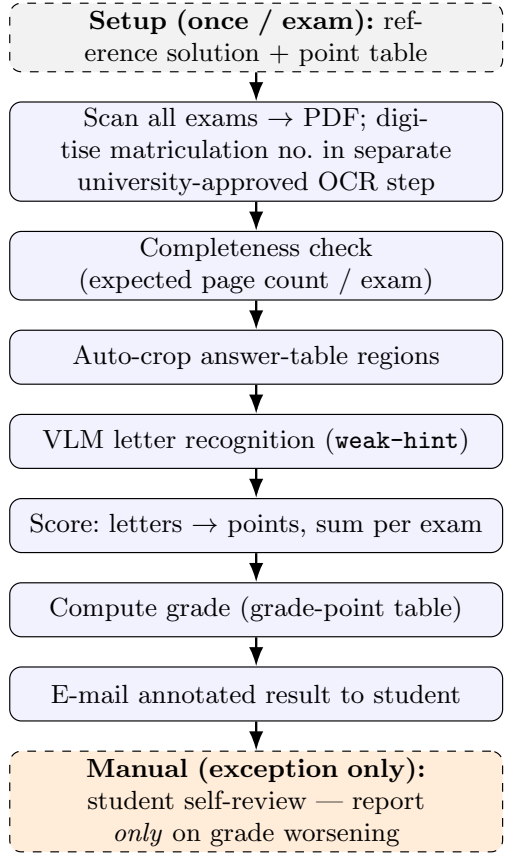

The recognition rate is now high enough to operate a fully \emph{unattended}
correction pipeline (Fig.~\ref{fig:pipeline}). After a one-time setup (reference
solution and per-question point values), the whole batch runs without human
intervention: all exams are scanned into a single PDF and the matriculation
number is digitised in a separate university-approved OCR step for
identification; this identifier crop is processed separately from the answer
recognition. Each exam is checked for completeness (expected page count); the
answer-table regions are cropped; the letters are recognised in
\texttt{weak-hint} mode; points are assigned and summed; the grade is computed
via the grade-point table; and an annotated result is e-mailed to each student.
The only manual step is student-triggered:
the e-mail asks students to respond \emph{only} if their grade would worsen
(a potential disadvantage), so human review is restricted to the few
grade-worsening cases (Table~\ref{tab:grades}).

\paragraph{Why this matters for large cohorts.}
Foundational courses routinely have several hundred examinees (our exam had
100 students; 61 exams were released for this study). Such cohorts are hard to
assess with fully digital exams --- which would need hundreds of synchronised,
supervised workstations and tolerate no system outage --- and prohibitively
tedious to grade manually, where tutors must read handwritten letters across
hundreds of sheets. Because every step of the pipeline is automatic, the human
effort no longer grows with the number of exams but only with the small number
of complaints. We validate the recognition stage on 61 exams; since the
pipeline is unattended and per-exam independent, it scales to large cohorts as
an offline batch process. To our knowledge, this is the first demonstration of
a recognition rate sufficient for \emph{hands-off} correction of large
paper-based exam cohorts, turning a previously infeasible task into a routine
batch job.

\section{Threats to Validity}
The benchmark consists of 61 exams from a single institution and author;
broader validation across institutions and handwriting styles is future work.
The grade-relevance estimate is indicative (see footnote). The strongest
models are closed, commercial services; results depend on the specific model
snapshots (Table~\ref{tab:main}) and may drift over time, and per-request cost
and latency were not the focus here. Student identifiers are processed only for
result delivery: matriculation-number crops are digitised separately through a
university-approved OCR service and are never included in the answer-table
crops sent to commercial VLM providers. Answer-recognition calls are restricted
to cropped answer tables without names or matriculation numbers. Finally,
provider routing matters: for the newest Gemini model we observed schema-compliance failures via a
third-party router that disappeared using the native API; such transport
effects must be separated from genuine recognition errors.

\section{Conclusion}
We have shown that general-purpose vision--language foundation models close
the gap left by our previous detector-based system: they recognise handwritten
exam answers at \SI{\sim98}{\percent} accuracy and, unlike a trained detector,
handle semantically complex cases such as out-of-cell, crossed-out and cursive
answers. With a simple weak-hint prompt the false-negative rate drops below
\SI{1}{\percent}, and under a uniform exemplary grading table only three of the
61 exams would be graded worse, all caught by student self-review. Fully
automated exam correction with student feedback is therefore defensible,
offering a concrete simplification of a costly and error-prone manual task.
Crucially, because the correction pipeline is unattended and per-exam
independent, it scales to large paper-based cohorts (several hundred students)
that are impractical to examine fully digitally and prohibitive to grade
manually --- enabling, for the first time, hands-off correction at that scale.
Future work includes multi-institution validation, a field deployment that
quantifies the actual reduction in correction effort, and on-premise
open-weight models for data-protection-sensitive settings.

\section*{Declarations}

\noindent\textbf{Funding.} The author received no specific funding for this
study.

\smallskip\noindent\textbf{Competing interests.} The author declares no
competing interests.

\smallskip\noindent\textbf{Ethics approval and consent to participate.} The
data originate from a written university examination conducted by the author.
The students gave informed, voluntary consent for their handwritten characters
to be processed in anonymised form for research and development; this consent
was revocable at any time with effect for the future. All exam sheets were
anonymised --- names and matriculation numbers were removed --- so that no link
to a person, identity, or examination result is possible. In the operational
workflow, matriculation numbers serve only to return results to students; they
are digitised in a separate, university-approved OCR step and are never
included in the released data nor in the answer-table requests sent to
commercial VLM providers.

\smallskip\noindent\textbf{Consent for publication.} The consent explicitly
covers the use of anonymised examples for scientific documentation,
publication, and reproducibility of the results.

\smallskip\noindent\textbf{Data availability.} The anonymised data of the 61
exams (\num{3141} answer positions; only the letter answers, no names or
matriculation numbers) are deposited on Zenodo
(\url{https://doi.org/10.5281/zenodo.20575275}) under \emph{restricted access}
and are available from the author upon reasonable request for scientific
reproducibility.

\smallskip\noindent\textbf{Use of generative AI.} Generative-AI language tools
were used only to improve phrasing and readability. All content, experiments,
analyses, and conclusions were produced and verified by the author.

\bibliography{refs}

@incollection{grabowski2023icr,
  author    = {Grabowski, Hartwig},
  title     = {Intelligent Character Recognition of Handwritten Forms with Deep Neural Networks},
  booktitle = {Towards AI-Aided Invention and Innovation},
  series    = {IFIP Advances in Information and Communication Technology},
  publisher = {Springer Nature Switzerland},
  address   = {Cham},
  pages     = {81--94},
  year      = {2023},
  doi       = {10.1007/978-3-031-42532-5\_6},
  isbn      = {978-3-031-42531-8}
}

@misc{grabowski2026hybrid,
  author       = {Grabowski, Hartwig and Canz, Michael},
  title        = {Hybrid {E}-Assessment in Higher Education: Semi-Automated Grading of Paper-Based Written Examinations},
  year         = {2026},
  howpublished = {arXiv:2606.08855},
  doi          = {10.48550/arXiv.2606.08855}
}

@inproceedings{cohen2017emnist,
  author    = {Cohen, Gregory and Afshar, Saeed and Tapson, Jonathan and van Schaik, Andr{\'e}},
  title     = {{EMNIST}: Extending {MNIST} to handwritten letters},
  booktitle = {International Joint Conference on Neural Networks (IJCNN)},
  pages     = {2921--2926},
  year      = {2017},
  publisher = {IEEE},
  doi       = {10.1109/IJCNN.2017.7966217}
}

@article{bommasani2021foundation,
  author        = {Bommasani, Rishi and Hudson, Drew A. and Adeli, Ehsan and Altman, Russ and Arora, Simran and von Arx, Sydney and Bernstein, Michael S. and Bohg, Jeannette and Bosselut, Antoine and Brunskill, Emma and others},
  title         = {On the Opportunities and Risks of Foundation Models},
  journal       = {arXiv preprint arXiv:2108.07258},
  year          = {2021},
  doi           = {10.48550/arXiv.2108.07258}
}

@article{garrido2026htrsurvey,
  author  = {Garrido-Munoz, Carlos and Rios-Vila, Antonio and Calvo-Zaragoza, Jorge},
  title   = {Handwritten Text Recognition: A Survey},
  journal = {IEEE Transactions on Pattern Analysis and Machine Intelligence},
  volume  = {48},
  number  = {4},
  pages   = {4367--4387},
  year    = {2026},
  doi     = {10.1109/TPAMI.2025.3646002}
}

@inproceedings{li2023trocr,
  author    = {Li, Minghao and Lv, Tengchao and Chen, Jingye and Cui, Lei and Lu, Yijuan and Florencio, Dinei and Zhang, Cha and Li, Zhoujun and Wei, Furu},
  title     = {{TrOCR}: Transformer-Based Optical Character Recognition with Pre-trained Models},
  booktitle = {Proceedings of the AAAI Conference on Artificial Intelligence},
  volume    = {37},
  pages     = {13094--13102},
  year      = {2023},
  doi       = {10.1609/aaai.v37i11.26538}
}

@inproceedings{kim2022donut,
  author    = {Kim, Geewook and Hong, Teakgyu and Yim, Moonbin and Nam, Jeongyeon and Park, Jinyoung and Yim, Jinyeong and Hwang, Wonseok and Yun, Sangdoo and Han, Dongyoon and Park, Seunghyun},
  title     = {{OCR}-free Document Understanding Transformer},
  booktitle = {European Conference on Computer Vision (ECCV)},
  pages     = {498--517},
  year      = {2022},
  publisher = {Springer},
  doi       = {10.1007/978-3-031-19815-1_29}
}

@inproceedings{blecher2023nougat,
  author    = {Blecher, Lukas and Cucurull, Guillem and Scialom, Thomas and Stojnic, Robert},
  title     = {Nougat: Neural Optical Understanding for Academic Documents},
  booktitle = {International Conference on Learning Representations (ICLR)},
  year      = {2024},
  note      = {arXiv:2308.13418},
  doi       = {10.48550/arXiv.2308.13418}
}

@article{li2023blip2,
  author        = {Li, Junnan and Li, Dongxu and Savarese, Silvio and Hoi, Steven},
  title         = {{BLIP-2}: Bootstrapping Language-Image Pre-training with Frozen Image Encoders and Large Language Models},
  journal       = {arXiv preprint arXiv:2301.12597},
  year          = {2023},
  doi           = {10.48550/arXiv.2301.12597}
}

@inproceedings{liu2023llava,
  author    = {Liu, Haotian and Li, Chunyuan and Wu, Qingyang and Lee, Yong Jae},
  title     = {Visual Instruction Tuning},
  booktitle = {Advances in Neural Information Processing Systems},
  volume    = {36},
  pages     = {34892--34916},
  year      = {2023}
}

@article{yang2024ccocr,
  author        = {Yang, Zhibo and Tang, Jun and Li, Zhaohai and Wang, Pengfei and Wan, Jianqiang and Zhong, Humen and Liu, Xuejing and Yang, Mingkun and Wang, Peng and Bai, Shuai and Jin, Lianwen and Lin, Junyang},
  title         = {{CC-OCR}: A Comprehensive and Challenging {OCR} Benchmark for Evaluating Large Multimodal Models in Literacy},
  journal       = {arXiv preprint arXiv:2412.02210},
  year          = {2024},
  doi           = {10.48550/arXiv.2412.02210}
}

@article{bai2023qwen,
  author        = {Bai, Jinze and Bai, Shuai and Yang, Shusheng and Wang, Shijie and Tan, Sinan and Wang, Peng and Lin, Junyang and Zhou, Chang and Zhou, Jingren},
  title         = {{Qwen-VL}: A Versatile Vision-Language Model for Understanding, Localization, Text Reading, and Beyond},
  journal       = {arXiv preprint arXiv:2308.12966},
  year          = {2023},
  doi           = {10.48550/arXiv.2308.12966}
}

@article{liu2024ocrbench,
  author        = {Liu, Yuliang and Li, Zhang and Huang, Mingxin and Yang, Biao and Yu, Wenwen and Li, Chunyuan and Yin, Xucheng and Liu, Cheng-Lin and Jin, Lianwen and Bai, Xiang},
  title         = {{OCRBench}: On the Hidden Mystery of {OCR} in Large Multimodal Models},
  journal       = {Science China Information Sciences},
  volume        = {67},
  number        = {12},
  pages         = {220102},
  year          = {2024},
  doi           = {10.1007/s11432-024-4235-6}
}

@article{fadeeva2024onlinehandwriting,
  author        = {Fadeeva, Anastasiia and Schlattner, Philippe and Maksai, Andrii and Collier, Mark and Kokiopoulou, Efi and Berent, Jesse and Musat, Claudiu},
  title         = {Representing Online Handwriting for Recognition in Large Vision-Language Models},
  journal       = {arXiv preprint arXiv:2402.15307},
  year          = {2024},
  doi           = {10.48550/arXiv.2402.15307}
}

@article{crosilla2025llmhtr,
  author  = {Crosilla, Giorgia and Klic, Lukas and Colavizza, Giovanni},
  title   = {Benchmarking Large Language Models for Handwritten Text Recognition},
  journal = {Journal of Documentation},
  volume  = {81},
  number  = {7},
  pages   = {334--354},
  year    = {2025},
  doi     = {10.1108/JD-03-2025-0082}
}

@misc{yolov5,
  author       = {Jocher, Glenn and others},
  title        = {{YOLOv5} by {Ultralytics}},
  howpublished = {\url{https://github.com/ultralytics/yolov5}},
  year         = {2020},
  note         = {Accessed June 2026}
}

@misc{ultralytics,
  author       = {{Ultralytics}},
  title        = {{Ultralytics YOLO} (YOLO26)},
  howpublished = {\url{https://github.com/ultralytics/ultralytics}},
  year         = {2026},
  note         = {Accessed June 2026}
}

@misc{gemini31flashlite,
  author       = {{Google DeepMind}},
  title        = {{Gemini 3.1 Flash-Lite} Model Card},
  howpublished = {\url{https://deepmind.google/models/model-cards/gemini-3-1-flash-lite/}},
  year         = {2026},
  note         = {Accessed June 2026}
}

@misc{gemini35flash,
  author       = {{Google DeepMind}},
  title        = {{Gemini 3.5 Flash} Model Card},
  howpublished = {\url{https://deepmind.google/models/model-cards/gemini-3-5-flash/}},
  year         = {2026},
  note         = {Published 19 May 2026; accessed June 2026}
}

@misc{gpt52,
  author       = {{OpenAI}},
  title        = {Update to the {GPT-5} System Card: {GPT-5.2}},
  howpublished = {\url{https://openai.com/index/gpt-5-system-card-update-gpt-5-2/}},
  year         = {2026},
  note         = {Accessed June 2026}
}

@misc{grok43,
  author       = {{xAI}},
  title        = {{Grok 4.3} Model Documentation},
  howpublished = {\url{https://docs.x.ai/developers/models/grok-4.3}},
  year         = {2026},
  note         = {Accessed June 2026 (API release 6 May 2026)}
}

@article{qwen3vl,
  author       = {{Qwen Team}},
  title        = {{Qwen3-VL} Technical Report},
  journal      = {arXiv preprint arXiv:2511.21631},
  year         = {2025},
  doi          = {10.48550/arXiv.2511.21631}
}

\end{document}